\documentclass{article}

\PassOptionsToPackage{numbers,square}{natbib}
 \usepackage{neurips_2025}
 \usepackage{graphicx}
 \usepackage{subcaption} 
 \usepackage{adjustbox}
 \usepackage{booktabs,tabularx,array}
\usepackage{algorithm}
\usepackage{algpseudocode}

\usepackage{amsmath,amssymb,bm}
\usepackage[switch]{lineno}

\usepackage[utf8]{inputenc} 
\usepackage[T1]{fontenc}    
\usepackage{hyperref}       
\usepackage{url}            
\usepackage{booktabs}       
\usepackage{amsfonts}       
\usepackage{nicefrac}       
\usepackage{microtype}      
\usepackage{xcolor}         

\title{Towards Self-Supervised Foundation Models for Critical Care Time Series}

\author{%
  David S.~Hippocampus\thanks{Use footnote for providing further information
    about author (webpage, alternative address)---\emph{not} for acknowledging
    funding agencies.} \\
  Department of Computer Science\\
  Cranberry-Lemon University\\
  Pittsburgh, PA 15213 \\
  \texttt{hippo@cs.cranberry-lemon.edu} \\
}

\begin{document}

\maketitle

\begin{abstract}

Domain-specific foundation models for healthcare have expanded rapidly in recent years, yet foundation models for critical care time series remain relatively underexplored due to the limited size and availability of datasets. In this work, we introduce an early-stage pre-trained foundation model for critical care time-series based on the Bi-Axial Transformer (BAT), trained on pooled electronic health record datasets. We demonstrate effective transfer learning by fine-tuning the model on a dataset distinct from the training sources for mortality prediction, where it outperforms supervised baselines, particularly for small datasets ($<5,000$). These contributions highlight the potential of self-supervised foundation models for critical care times series to support generalizable and robust clinical applications in resource-limited settings.\\
\textbf{Code Availability}: \url{https://github.com/Katja-Jagd/YAIB}

\end{abstract}

\section{Introduction and related works}
\label{Introduction}
Foundation models built on Transformers \citep{vaswani2017attention} have achieved remarkable success across several domains, including natural language processing \cite{devlin2019bertpretrainingdeepbidirectional, radford2018improving} and computer vision \cite{dosovitskiy2021imageworth16x16words, liu2021swintransformerhierarchicalvision, touvron2021trainingdataefficientimagetransformers}. Despite this success, general-purpose foundation models often perform poorly on healthcare applications, which are impeded by the complexities of medical data and the scarcity of publicly available labeled datasets \cite{khan2025a}. Consequently, the development of healthcare-specific foundation models has accelerated rapidly since 2018, targeting diverse applications such as clinical natural language processing \cite{zeming2024a, xie2024a}, medical imaging \cite{ma2024a, cox2024a}, omics analysis \cite{zhou2024a, Celaj2023.09.20.558508}, video and audio interpretation \cite{luo2023a, wang2023a} and multi-modality \cite{zhang2025a, yuan2025a}. Foundation models for Electronic Health Records (EHRs) have primarily focused on structured EHR data, such as modelling and predicting ICD-10 codes \cite{wornow2023aehrshot, steinberg2023motortimetoeventfoundationmodel}. Although these models can scale to large patient populations, they remain limited in their ability to capture physiological patterns \cite{Burger2025.07.25.25331635}.  One area that remains relatively underexplored is the development of foundation models for critical care time-series data \cite{burger2024TowardsFoundationModels}.
Models trained in this domain are often hindered by issues of reproducibility \cite{water2024aYAIB}, limited by simple learning paradigms that rely on one supervised task \cite{horn2020aseft, che2018agrud, luo2025hipatch} or are trained on a small, homogeneous dataset \cite{harutyunyan2019amultitask, yèche2022hirid, fallahpour2024amamba}. It has been demonstrated that these models do not transfer well to new clinical settings \cite{burger2024TowardsFoundationModels}. Recent work from \citet{Burger2025.07.25.25331635} are the first to combine pooling of several critical care time-series datasets with self-supervised pretraining, and thereby delivering a critical care time series foundation model, ICareFM \cite{Burger2025.07.25.25331635}. Both the model and code are currently not open-source. 

To push critical care time-series data towards foundation model development, we modify the Bi-axial Transformer (BAT) architecture, presented in the work of \citet{devries2025biaxialtransformersaddressingincreasing}, for self-supervised pre-training and conduct all experiments within the Yet Another ICU Benchmark (YAIB) framework \cite{water2024aYAIB}, ensuring transparency, reproducibility and pooling of datasets. Our contributions are as follows: 1) We release the first ICU-specific model for foundational capabilities with an open-source, reproducible repository, 2) we demonstrate its ability to transfer effectively to an unseen dataset distinct from the training sources and a new downstream clinical task, outperforming supervised baselines, and 3) we show that it performs especially well in low-data regimes, highlighting the potential of self-supervised pre-training for resource-limited clinical settings that may not achieve robust performance from supervised models.

\section{Methodology}
\label{Methodology}

Let \(\mathcal{D} := \bigcup^K_{k=1} \mathcal{D}_k\), where a critical care dataset \(\mathcal{D}_k\) is defined as a set of tuples, \(\left\{ \left( \mathbf{X}_i,\ \mathbf{p}_i,\ \mathbf{h}_i \right) \right\}_{i=1}^{N_k}\). Every patient \(i \in N_k\) is represented with a multivariate time series of measurements \(\mathbf{X}_i \in \mathbb{R}^{T_i \times D}\) from monitoring devices and laboratory tests, with \(K\) datasets containing the same measurement types \(D\). As each \(\mathbf{X}_i\) is irregularly sampled across both \(t\) and \(d\), we also record measurement times \(\mathbf{h}_i\in \mathbb{R}^{T_i}\). The time series is supplemented with patient demographics \(\mathbf{p}_i\) which are fixed for the duration of $T$.


In critical care applications, it is often of interest to predict a class label, such as $y_i \in \{0,1\}$ for mortality status, given $(\mathbf{X}_i, \mathbf{p}_i, \mathbf{h}_i)$. Popular approaches split a single dataset $\mathcal{D}_1$ into train/test/validation subsets, and perform supervised training to predict $y_i$. We hypothesize that combining and pretraining on additional datasets $\{\mathcal{D}_2, ... \mathcal{D}_K\}$, in a self-supervised fashion will result in a model that learns richer, more generalized representations of sensor identities and their measurement distributions.

\subsection{Pretraining and Fine-tuning}
\label{pretrianing}

We begin with self-supervised pretraining with forecasting as the prediction task. Specifically, \(\mathbf{X}_i\) sampled from the auxiliary datasets $\{\mathcal{D}_2, ... \mathcal{D}_K\}$ are split into an observation window \(\mathbf{X}_i^{\mathrm{obs}} \in \mathbb{R}^{T^{\mathrm{obs}} \times D}\) and a forecasting window \(\mathbf{X}_i^{\mathrm{for}} \in \mathbb{R}^{T^{\mathrm{for}} \times D}\). The learning objective is to predict \(\mathbf{X}_i^{\mathrm{for}}\) given \((\mathbf{X}_i^{\mathrm{obs}}, \mathbf{p}_i, \mathbf{h}_i^{\mathrm{obs}})\). Due to the irregular sampling across $h_i$ and sparsity in $X_i$, the forecasting must be sample-specific. To handle sparsity and ensure temporal alignment, we leverage a masked loss:

\begin{linenomath*}
\begin{equation*}
\mathcal{L}^{\text{Pre}}
= \frac{1}{\sum_{k=2}^{K} N_k}\ \sum_{k=2}^{K}\sum_{i=1}^{N_k}
\left\lVert
  \mathbf{M}_i^{\mathrm{for}} \odot
  \bigl(\hat{\mathbf{X}}_i^{\mathrm{for}} - \mathbf{X}_i^{\mathrm{for}}\bigr)
\right\rVert_F^{2}
\label{eq:pretrain}
\end{equation*}
\end{linenomath*}

where the mask 
\(\mathbf{M}_i^{\mathrm{for}}\) ensures that only observed values in the forecasting window contribute to the loss. 

We then fine-tune and test on \(\mathcal{D}_1\) with supervised learning, where the objective is our initial goal of predicting the mortality status \(y_i \in \{0,1\}\) as a binary classification task. We used the binary cross-entropy loss:

\begin{linenomath*} 
\begin{equation*}
\mathcal{L}^{\text{Fine}}
= -\frac{1}{N_1}\sum_{i=1}^{N_1}
\Bigl[y_i \log(\hat{y}_i) + (1-y_i)\log(1-\hat{y}_i)\Bigr]
\label{eq:finetune}
\end{equation*}
\end{linenomath*}


\subsection{Bi-axial Transformer (BAT)}
\label{Bi-axial Transformer (BAT)}

The Bi-Axial Transformer (BAT) \citep{devries2025biaxialtransformersaddressingincreasing} is a generalization of the Axial Transformer \citep{ho2019axial} with parallel axial attention along each dimension. BAT attends to both the temporal and clinical feature axes through axial attention mechanisms, and explicitly accounts for missing values. The model architecture is described in detail in Appendix~\ref{BAT architecture}. BAT shows promising results in prediction tasks for clinical irregular multivariate time series data with state-of-the-art performance for sepsis classification, and competitive performance on mortality prediction when compared to several models.  Investigation into model attention maps revealed evidence of BAT learning from informative missingness, and it showed an increased robustness to sparsity in comparison to other Transformer models. These properties make BAT a strong candidate for modeling clinical multivariate time series. BAT's prediction head was adapted in this work to support both binary classification, outputting \( \hat{y} \in \{0, 1\} \), and forecasting, outputting \( \hat{\mathbf{X}}^{\mathrm{for}} \in \mathbb{R}^{T^{\mathrm{for}} \times D} \), as illustrated in Figure~\ref{fig:bat_architecture}.

\subsection{Yet another ICU benchmark (YAIB)}
\label{Yet another ICU benchmark (YAIB)}

Yet another ICU benchmark (YAIB) provided by \citet{water2024aYAIB} is a modular, end-to-end framework supporting transparent benchmarking for clinical machine learning on ICU data (Appendix~\ref{Yet another ICU benchmark framework}). YAIB supports several publicly available EHR datasets and enables cohort selection, harmonization, and a variety of supervised learning tasks, with several implemented machine learning and deep learning models. This work has extended the YAIB framework to contain a self-supervised learning module, detailed in Appendix~\ref{Implementation of self-supervised learning}, as well as the previously described BAT architecture. Building on YAIB’s emphasis on transparent and reproducible benchmarking, we use the framework to evaluate our self-supervised models in a controlled setting. 

\section{Experiments}
\label{Experiments}

All experiments were performed using the Medical Information Mart for Intensive Care Database versions III and IV (MIMIC-III \& IV, \cite{johnson2016a, johnson2023a}) and The eICU Collaborative Research Database (eICU \cite{pollard2018a}). Further information about the three datasets and their preprocessing can be found in Appendix~\ref{Datasets, cohort definitions and preprocessing}.
A t-SNE \cite{van2008atsne} analysis of the preprocessed and harmonized datasets was performed, and Figure~\ref{tsne} indicates that they occupy a similar overall distribution. Despite the overlap in data distribution, we reproduce similar findings to previous studies \cite{burger2024TowardsFoundationModels, rockenschaub2024aTheImpactOfMulti-institution} showing that training models on one dataset does not transfer effectively to new ones during inference (Figure~\ref{SCT}). Hyperparameters for all models can be found in Appendix~\ref{Hyperparameters}.

\subsection{Baseline Comparison}
BAT was pre-trained with self-supervised forecasting on a pooled version of the datasets, with one held-out for fine-tuning. Pre-training followed the cross-validation setup described in Appendix~\ref{Data splits}, and details on the models can be found in Appendix~\ref{Pre-trained models}. The pre-trained models were evaluated on the ability to be fine-tuned and generalize to the held-out dataset for mortality prediction. To test model robustness, fine-tuning was performed on varying subsets of the held-out data, with training set sizes ranging from 100 to 9,506 samples. 

\begin{figure}[!htbp]
  \centering
  \begin{minipage}[b]{0.46\textwidth}
    \centering
    \includegraphics[width=\textwidth]{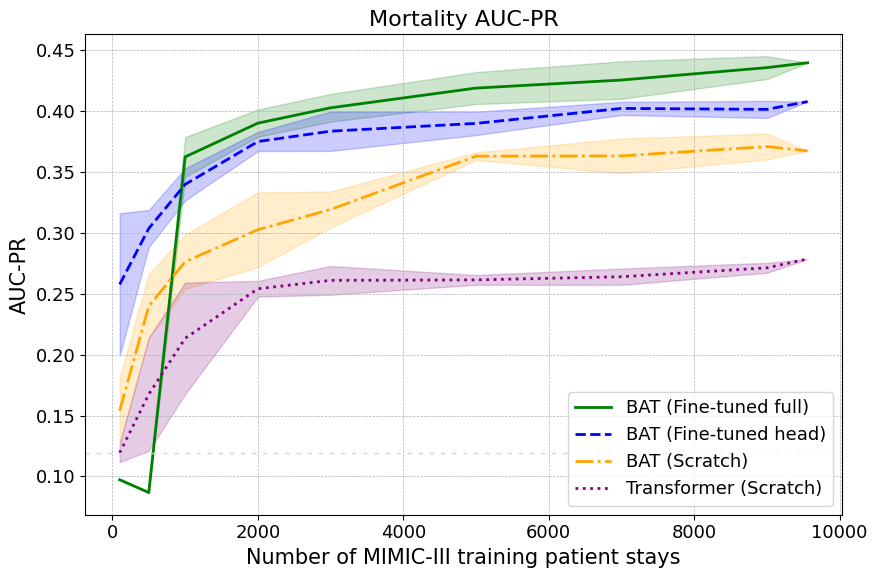}
    \\[0.3em] 
    \hspace*{2em}\textbf{(a)} 
  \end{minipage}
  \hfill
  \begin{minipage}[b]{0.46\textwidth}
    \centering
    \includegraphics[width=\textwidth]{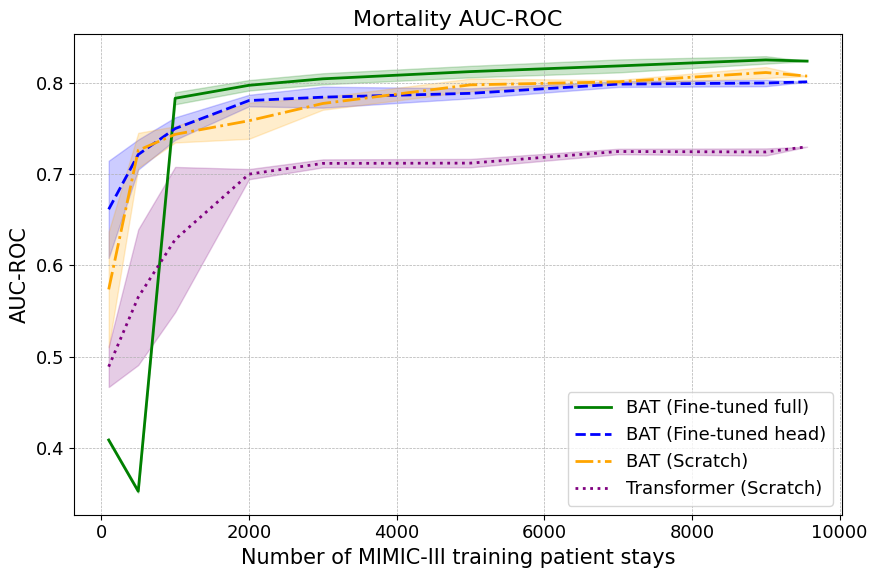}
    \\[0.3em]
    \hspace*{2em}\textbf{(b)} 
  \end{minipage}
  \vspace{1em} 
  \caption[Fine-tuning Results of Foundation Model]{Performance of fine-tuned model (pre-trained on eICU + MIMIC-IV) and supervised models trained from scratch on MIMIC-III. \textbf{(a)} AUC-PR (grey dashed line indicates the positive class prevalence) and \textbf{(b)} AUC-ROC across training set sizes ranging from 100 to 9{,}506 samples. Models include BAT full \& head fine-tuning, and models trained from scratch: BAT and Transformer.}
  \label{fig:fine-tuning}
\end{figure}

Figure~\ref{fig:fine-tuning} illustrates the results of pre-training on the pooled dataset  consisting of eICU \& MIMIC-IV and fine-tuning on MIMIC-III. The mean and standard deviations of AUC-PR and AUC-ROC over all training set sizes reflects five random subsamples of the training sets with preserved class imbalance. The pre-trained model was compared to two baseline models trained from scratch in a supervised manner: one using the same architecture (BAT) and another based on a vanilla Transformer that does not natively handle missing values and instead relies on mean-imputed data. The AUC-PR performances, and partially the AUC-ROC performance, revealed that the pre-trained model outperformed the baseline models on all training set sizes $>500$ samples, but most significantly on dataset sizes $<5000$ samples. Furthermore, fine-tuning only the binary classification head of the pre-trained model yields close to similar performance when compared to full-model fine-tuning. Similar results were seen when MIMIC-IV and eICU were held out (Table~\ref{tab:fine-tuning_PR_highlighted_results}). In these two cases, the BAT model trained from scratch outperformed the fine-tuned version where only the classification head was fine-tuned, when enough labeled data was available ($\geq 3000$ and $\geq 5000$ samples, respectively). This suggests that pre-training on pooled eICU+MIMIC-IV data leads to more robust representations compared to the other pooled pre-training datasets. Results for held-out test sets of additional sizes are provided in Appendix~\ref{Fine-tuning results}, and show similar performance.
\begin{table}[!htbp]
\centering
\setlength{\tabcolsep}{6pt}
\renewcommand{\arraystretch}{1.1}
\caption{Average model performance, (AUC-PR $\pm$ sd) across multiple dataset sizes from MIMIC-III, MIMIC-IV, and eICU. The pre-trained BAT models are fine-tuned and the baseline models are trained from scratch on the subsets. Highest performance for each dataset size is in bold, second highest is underlined.}
\vspace{1em} 
\label{tab:fine-tuning_PR_highlighted_results}

\begin{adjustbox}{width=\linewidth,center}
\begin{tabular}{l r c c c c}
\hline
\noalign{\vskip 2pt}
\shortstack{Fine-tuning/\\Training dataset} & \shortstack{Dataset\\size}
& \shortstack{BAT\\(Fine-tuned full)}
& \shortstack{BAT\\(Fine-tuned head)}
& \shortstack{BAT\\(Scratch)}
& \shortstack{Transformer\\(Scratch)} \\
\hline
          & 1000 & \textbf{36.24 $\pm$ 1.63} & \underline{33.98 $\pm$ 1.32} & 27.63 $\pm$ 2.23 & 21.34 $\pm$ 4.58 \\
\multicolumn{1}{c}{MIMIC-III} & 5000 & \textbf{41.89 $\pm$ 1.31} & \underline{38.99 $\pm$ 0.96} & 36.30 $\pm$ 0.31 & 26.14 $\pm$ 0.40 \\
          & 9000 & \textbf{43.57 $\pm$ 0.94} & \underline{40.14 $\pm$ 0.70} & 37.09 $\pm$ 1.06 & 27.13 $\pm$ 0.41 \\
\hline
          & 1000 & \textbf{28.98 $\pm$ 0.85} & \underline{26.97 $\pm$ 1.71} & 26.12 $\pm$ 1.95 & 13.06 $\pm$ 1.56 \\
\multicolumn{1}{c}{MIMIC-IV} & 5000 & \textbf{38.10 $\pm$ 1.36} & 31.31 $\pm$ 1.17 & \underline{34.97 $\pm$ 1.03} & 18.00 $\pm$ 1.11 \\
          & 9000 & \textbf{38.75 $\pm$ 0.53} & 32.19 $\pm$ 1.00 & \underline{38.41 $\pm$ 0.88} & 18.91 $\pm$ 1.36 \\
\hline 
          & 1000 & \textbf{28.37 $\pm$ 1.13} & \underline{25.39 $\pm$ 1.56} & 20.86 $\pm$ 3.31 &  6.58 $\pm$ 4.00 \\
\multicolumn{1}{c}{eICU}& 5000 & \textbf{33.89 $\pm$ 0.49} & 29.09 $\pm$ 0.62 & \underline{30.13 $\pm$ 1.37} & 14.41 $\pm$ 0.42 \\
          & 9000 & \textbf{35.20 $\pm$ 0.87} & 29.53 $\pm$ 0.95 & \underline{31.59 $\pm$ 0.94} & 14.61 $\pm$ 1.06 \\
\hline
\end{tabular}
\end{adjustbox}
\end{table}
\section{Discussion}
\label{Discussion}
We evaluate BAT as a model architecture for a critical care time-series foundation model and find that it outperforms initial supervised baselines on held-out datasets distinct from the training sources for mortality prediction, with training set sizes up to ~10,000 samples. Fine-tuning only the binary classification head of the pre-trained model achieves performance comparable to full-model fine-tuning, suggesting that the learned embeddings are both informative and transferable to downstream tasks on unseen datasets. This effect was most pronounced for the model pretrained on the largest dataset (277K samples; see Appendix ~\ref{Pre-trained models}), MIMIC-IV + eICU, and fine-tuned on MIMIC-III. By contrast, the other two models with smaller training set sizes were outperformed by the equivalent baseline models trained from scratch when only the classification head was fine-tuned and sufficient labeled data was available. These findings indicate that large and diverse pre-training datasets are crucial for learning representations that generalize and transfer effectively across datasets. Together, these results highlight the value of such a model in clinical settings where labeled data and computational resources are limited. Overall, this work demonstrates the feasibility and benefits of training foundation models for critical care time-series data within a transparent, reproducible framework \cite{water2024aYAIB}, extended for self-supervised pre-training to enable transfer learning and more robust, generalizable models. 

\subsection{Limitations}
Our experiments were limited to MIMIC-III, MIMIC-IV, and eICU. To reach a training set size that is competitive with other foundation models, future work will need to incorporate additional datasets, such as those presented in \citet{burger2024TowardsFoundationModels}. Compared with fields like natural language processing, the number and size of publicly available ICU and critical care datasets remain very limited. Expanding the training data may therefore also require incorporating time-series datasets from other domains, such as weather \cite{bgcjena_weather} or electricity consumption \cite{trindade2015electricity}. This would allow us to assess whether exposure to broader time-series distributions improves model performance, or if domain-specific data is necessary given the sparse and irregular nature of critical care records, as suggested by other healthcare foundation models \cite{khan2025a}. 

\bibliographystyle{unsrtnat}
\bibliography{bibliography_neurips_2025}

\appendix
\newpage
\section{Appendix}
\label{Appendix}
\subsection{BAT architecture}
\label{BAT architecture}
\begin{figure}[htbp]
  \centering
  \vspace{1em} 
  \includegraphics[width=0.95\linewidth]{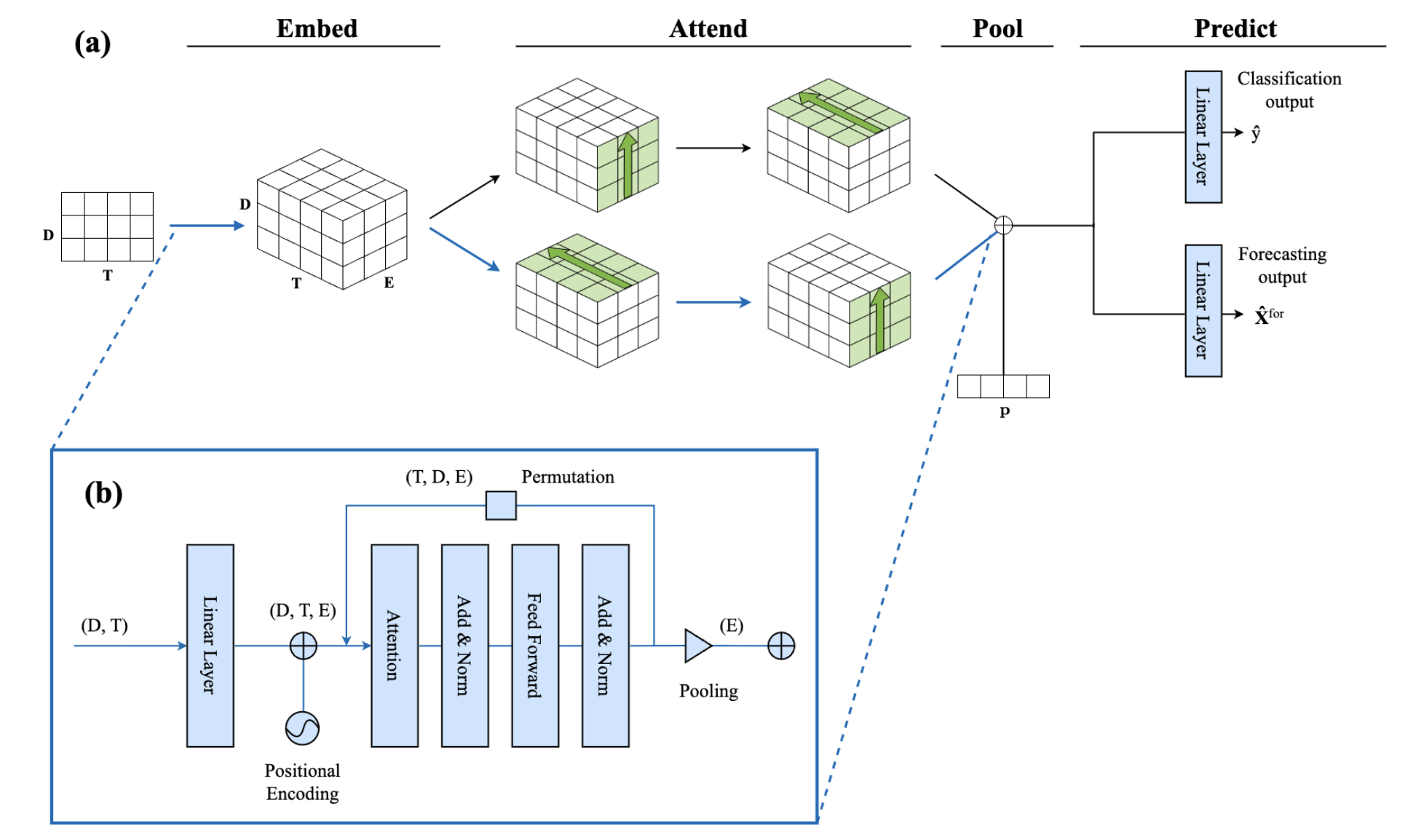}
  \vspace{1em} 
  \caption[Bi-Axial Transformer Architecture]{Overview of the Bi-Axial Transformer (BAT) architecture. (a) Shows the full model architecture and data representation, and (b) Shows an attention track indicated by the blue arrows. $D$ is the number of time-varying features, $T$ the number of time steps, $E$ the embedding size, and $P$ the static features. The model supports both binary classification and forecasting via separate prediction heads. This figure is adapted from Figure~2 in \citet{devries2025biaxialtransformersaddressingincreasing}.} 
  \vspace{1em} 
  \label{fig:bat_architecture}
\end{figure}

One Transformer-based model designed for irregular multivariate time series data is Bi-Axial Transformer (BAT) introduced in the work of \citet{devries2025biaxialtransformersaddressingincreasing}. Unlike classical Transformers that use an encoder-decoder structure for sequence generation, BAT relies solely on encoder-style self-attention to model dependencies across both the temporal and feature/sensor dimensions. This axial attention structure allows BAT to capture rich interactions within and across time steps and modalities. To address common challenges in irregular multivariate time series data, such as missing values and heterogeneous input types, BAT incorporates missingness indicators directly into its input embeddings. Figure~\ref{fig:bat_architecture} (Adapted from Figure~2 in \citet{devries2025biaxialtransformersaddressingincreasing}) illustrates the overall architecture of BAT. The input to the model is a multivariate time series of shape \( D \times T \), where \( D \) is the number of features and \( T \) is the number of time steps. Each input is embedded using a combination of the observed value (including missingness), a learned feature identity embedding, and a continuous time-based positional encoding. This results in an embedded tensor of shape \( D \times T \times E \), where \( E \) is the embedding dimension. The embedded input is then processed by bi-axial attention layers, which apply self-attention separately along the time and feature dimensions. The output from the attention layers is subsequently pooled and concatenated with the static features, \( P \), which include non-time-varying demographic variables such as age and sex. The BAT model was adapted in this work to have two different prediction heads: one for a supervised learning setup via binary classification, predicting \( \hat{y} \in \{0, 1\} \), and another for a self-supervised learning setup via forecasting, predicting \( \hat{\mathbf{X}}^{\mathrm{for}} \in \mathbb{R}^{T^{\mathrm{for}} \times D} \).

\newpage
\subsection{Yet another ICU benchmark framework}
\label{Yet another ICU benchmark framework}
\begin{figure}[htbp]
  \centering
  \vspace{1em} 
  \includegraphics[width=1\linewidth]{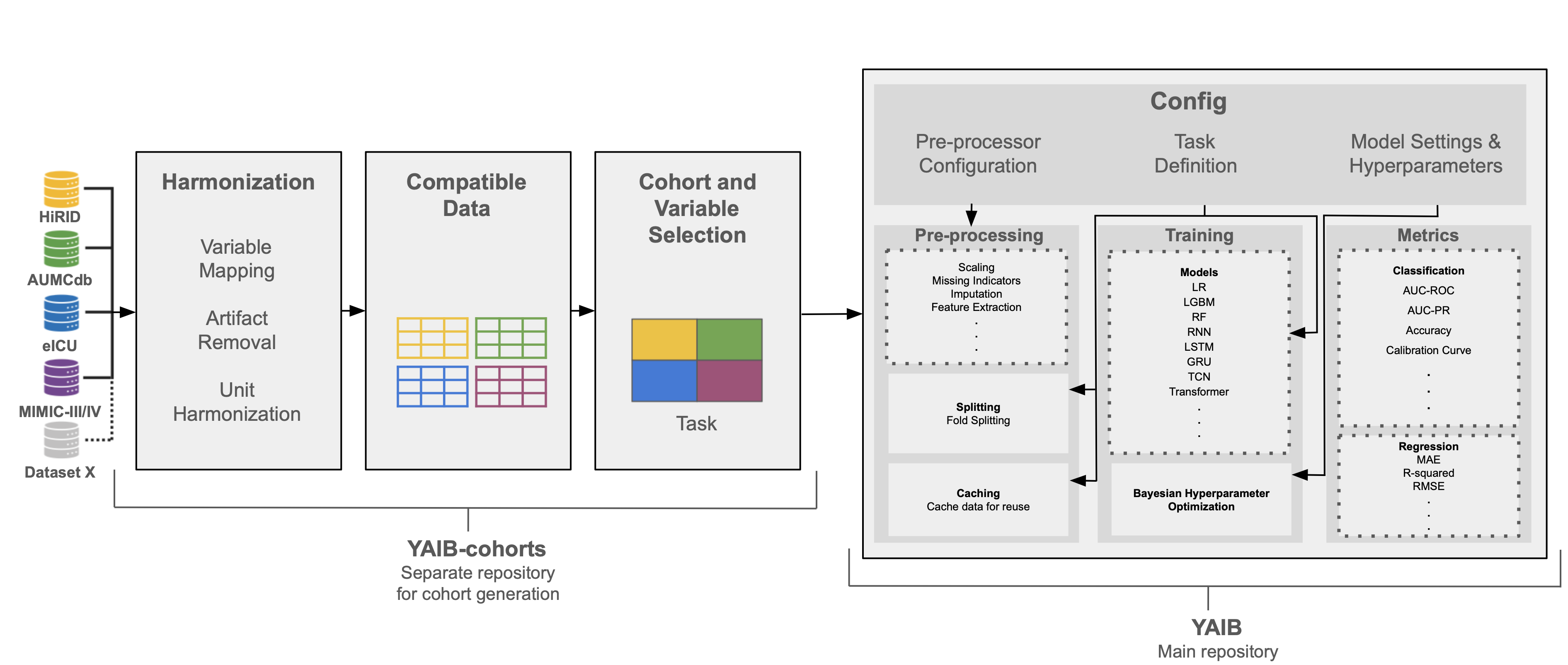}
  \vspace{1em} 
  \caption[Yet Another ICU Benchmark Pipeline]{Overview of the Yet Another ICU Benchmark (YAIB) pipeline. The left side illustrates the creation of harmonized ICU cohorts, implemented in a separate repository, \texttt{YAIB-cohorts} \cite{yaib_cohorts}. The right side represents the machine learning component of the pipeline, contained in the main \texttt{YAIB} repository \cite{yaib_main}, which covers preprocessing, model training, and evaluation. Dotted-line components indicate extensible modules that follow a standardized interface. This figure is inspired from Figure~1 in \citet{water2024aYAIB}.}
  \vspace{1em} 
  \label{fig:yaib}
\end{figure}

The Yet Another ICU Benchmark (YAIB) framework \cite{water2024aYAIB} provides a modular, end-to-end solution for clinical machine learning on ICU data, explicitly designed to address key limitations regarding reproducibility in the field. An illustration of the framework is shown in Figure~\ref{fig:yaib} (Inspired by Figure~1 in \citet{water2024aYAIB}). It consists of two repositories: \texttt{YAIB-cohorts} \cite{yaib_cohorts}, which handles dataset harmonization and cohort construction, and the main \texttt{YAIB} repository \cite{yaib_main}, which manages model training and evaluation.
The \texttt{YAIB-cohorts} repository builds on the open-source \texttt{R} package \texttt{ricu} \cite{bennett2023aricu} to harmonize multiple ICU datasets using a unified, concept-based abstraction of clinical variables. It supports five publicly available EHR datasets (MIMIC-III, MIMIC-IV, eICU, HiRID, AUMCdb) and maps their contents into a common structure with consistent semantic definitions and temporal alignment, and supports integration of new datasets. This harmonization enables standardized cohort construction, label definition, and data extraction across datasets, facilitating multi-center analyses and reproducible experimental setups. The main \texttt{YAIB} repository then provides the downstream machine learning pipeline for supervised modeling, including pre-processing, feature extraction, and evaluation. We entended the framework to support self-supervised pretraining, described in detail in Appendix~\ref{Implementation of self-supervised learning}, as well as add the BAT architecture to the selection of models. 
The models in this work show lower performance on the mortality prediction task compared to similar studies \cite{tipirneni2022astrats, patel2024aemit, labach2023aduett}, particularly in class sensitive metric, AUC-PR. As \citet{water2024aYAIB} notes, variations in preprocessing pipelines, task and cohort definitions etc. and limited transparency hinder reproducibility, especially when code is unavailable. In contrast to most prior work that experiment with performance-boosting strategies, e.g. upsampling of the minority class, our main focus of this work has been on demonstrating the transfer learning potential of robust self-supervised models rather than surpassing state-of-the-art results. We used a weighted loss available in the YAIB framework to handle the high class imbalance, but did not investigate further  performance-boosting strategies. This work benchmarks pre-trained models against two baselines, aiming to provide a comprehensive and fair comparison throughout the entire machine learning pipeline. 

\newpage
\subsection{Implementation of self-supervised learning}
\label{Implementation of self-supervised learning}
The self-supervised learning objective is implemented as a forecasting task, where future values are predicted based on past values. The approach in this work is inspired by the dynamic sampling method proposed by \citet{tipirneni2022astrats}, in which observation and forecasting windows are dynamically constructed during batch loading. A set of constraints is introduced to govern the selection of observation and forecasting windows. These constraints are designed to ensure both sufficient historical context and generating a valid forecasting window, thereby improving the overall quality and consistency of the training data. The overall goal is to select a valid time index where the observation and forecasting window are split. This work performs the search for the valid time index in a batch-wise manner. For each batch element, a patient is randomly sampled and candidate time indices are identified. These indices are filtered based on the following constraints:
\begin{enumerate}
    \item \textbf{Sparsity check:} The selected index must correspond to a time point where there is at least one observed value in the observation window. 
    \item \textbf{Minimum Observation Length:} The index must be at least $L$ time steps into the time-series to ensure sufficient historical context. For this work, $L = 12$ hours.
    \item \textbf{Forecasting Window Availability:} The index must allow room for a complete forecasting window of length $H$. For this work, $H = 2$ hours.
\end{enumerate}
If valid time indices are found, one is randomly selected as the split point between the observation window and the forecasting window. If no valid index is found, a new patient is sampled and the process is repeated. This approach ensures that the model is trained on different points within each hospital stay, with varying observation window lengths, effectively exposing it to diverse temporal contexts and clinical information.The sampling method are defined in Algorithm 1.  
\begin{algorithm}
\label{SSL}
\caption{Dynamic sampling of observation and forecasting window during batch loading}
\begin{algorithmic}[1]
\Function{Call}{batch}
    \State $(data, mask) \gets \textsc{LoadBatch}(batch)$
    \State $(B, C, T) \gets \textsc{Shape}(data)$
    \State $t_1 \gets \text{None}, \; tries \gets 0, \; max\_tries \gets B$

    \While{$t_1 = \text{None}$ \textbf{and} $tries < max\_tries$}
        \State $i \gets \textsc{RandomInteger}(0, B-1)$
        \State $valid\_times \gets \{t \mid mask[i, t] = \text{True}\}$

        \State $valid\_times \gets \{t \in valid\_times \mid t \geq L\}$
        \If{$valid\_times = \emptyset$}
            \State $tries \gets tries + 1$
            \State \textbf{continue}
        \EndIf

        \State $max\_index \gets \max(valid\_times)$
        \State $valid\_times \gets \{t \in valid\_times \mid t \leq max\_index - H\}$
        \If{$valid\_times = \emptyset$}
            \State $tries \gets tries + 1$
            \State \textbf{continue}
        \EndIf

        \State $t_1 \gets \textsc{RandomChoice}(valid\_times)$
    \EndWhile

    \If{$t_1 = \text{None}$}
        \State \textbf{raise} Error(``No valid index found in batch'')
    \EndIf

    \State $t_0 \gets \max(0, \; t_1 - max\_obs)$
    \State $t_2 \gets t_1 + forecast\_horizon$

    \State $X_{obs} \gets data[:, :, t_0 : t_1]$
    \State $M_{obs} \gets mask[:, :, t_0 : t_1]$

    \State $X_{for} \gets data[:, :, t_1 : t_2]$
    \State $M_{for} \gets mask[:, :, t_1 : t_2]$

    \State \Return $(X_{obs}, M_{obs}, X_{for}, M_{for})$
\EndFunction
\end{algorithmic}
\end{algorithm}
\newpage
\subsection{Datasets, cohort definitions and preprocessing}
\label{Datasets, cohort definitions and preprocessing}
The three publicly available ICU datasets used in this work are MIMIC-III \cite{johnson2016a}, MIMIC-IV \cite{johnson2023a}, and the eICU \cite{pollard2018a}. These datasets contain structured clinical data from ICU stays and are widely used in the development of machine learning models for critical care. Table~\ref{tab:dataset_stats} summarizes key statistics for the three datasets. As this study involves pooling of the datasets into one larger dataset, it is important to avoid patient overlap between them, as this could lead to data leakage in downstream experiments. Therefore, a filtered subset of the original MIMIC-III dataset that excludes all patients also found in MIMIC-IV is used. The overlap exists because MIMIC-IV was developed to extend MIMIC-III by including more recent and higher-resolution data. The subset of the original MIMIC-III dataset used in this work is called MIMIC-III Clinical Database CareVue subset \cite{johnson2022mimicsubset}, which limits MIMIC-III to records from 2001–2008 thereby excluding patient stays in the overlapping time period. Furthermore, a second filter was applied to remove any individual who also appears in MIMIC-IV, as some patients had an earlier stay recorded in MIMIC-III and a later hospital stay captured in MIMIC-IV. 
\begin{table}[htbp]
\centering
\caption[Dataset Statistics]{Dataset statistics for MIMIC-III, MIMIC-IV, and eICU. 
BIDMC = Beth Israel Deaconess Medical Center. *The subset of the MIMIC-III data 
used in this work is the MIMIC-III CareVue subset \cite{johnson2022mimicsubset}; 
values in parentheses represent statistics from the full MIMIC-III dataset.}
\vspace{1em} 
\label{tab:dataset_stats}
\renewcommand{\arraystretch}{1.5}
\begin{tabularx}{\textwidth}{l p{2.1cm} p{2.0cm} p{1.5cm} p{2.1cm} l}
\hline
\textbf{Dataset} & \textbf{Admissions} & \textbf{Collection} & \textbf{Origin} & \textbf{Hospital} & \textbf{Mortality} \\
                 &                     & \textbf{period}     &                 &                   & \textbf{(Positive class)} \\
\hline
MIMIC-III* & 27k* (40k) & 2001--2008* (2001--2012) & U.S. & BIDMC & 11.9\% \\
MIMIC-IV   & 73k        & 2008--2019                & U.S. & BIDMC & 7.3\%  \\
eICU       & 200k       & 2014--2015                & U.S. & 208 hospitals across the U.S. & 5.5\%  \\
\hline
\end{tabularx}
\end{table}
Feature selection was performed by \citet{water2024aYAIB} based on availability across all benchmarked datasets available in YAIB \cite{yaib_main}, with an emphasis on consistency to support cross-dataset experiments. A total of 52 clinical features were used as input for model development, consisting of 4 static and 48 time-varying variables. A full list of the features used and their units are provided in Table~\ref{tab:features}.

Patient cohorts were constructed using the \texttt{YAIB-cohorts} repository \cite{yaib_cohorts}. Data were temporally aligned and resampled at one-hour resolution. Uniform exclusion criteria were applied across all datasets and tasks: \textbf{(1)} invalid ICU stay timing (e.g., negative length of stay), \textbf{(2)} ICU stay < 6 hours, \textbf{(3)} fewer than four valid time points, \textbf{(4)} measurement gaps > 12 hours, and \textbf{(5)} age < 18 years at ICU admission. These filters preceded task-specific cohort definitions.
For mortality classification, the input window was the first 24 hours post-ICU admission; stays < 30 hours were excluded to prevent causal leakage \cite{water2024aYAIB}. The outcome label was mortality = 1 if the patient died during the same hospital admission. 
The YAIB framework requires task-specific definitions with inclusion/exclusion criteria and label generation, making it incompatible with self-supervised learning where labels are not used. Since YAIB cannot generate unlabeled cohorts, we selected the Length of Stay task, which imposed no additional exclusions beyond the five main criteria mentioned, thereby maximizing size of the pre-training dataset. Length of stay labels were still generated as a consequence of YAIB’s supervised design but were simply discarded resulting in a unlabeled cohort generation. This unlabeled cohort was the patient cohort used for pre-training. This approach maintained full compatibility with the existing framework while avoiding the need for significant changes to its design.  

Preprocessing was carried out using YAIB’s main repository \texttt{YAIB} \cite{yaib_main}, applied after cohort extraction. This included the addition of missingness indicators, forward-fill imputation within each ICU stay, and mean imputation for values without prior observations, using statistics computed from the training set to avoid data leakage. All features were standardized to zero mean and unit variance based on training split statistics. Data were split into training, validation, and test sets using cross-validation folds defined during preprocessing, ensuring that preprocessing steps such as scaling and imputation were fitted only on the training data. All preprocessing decisions followed the default setup specified by the \texttt{base\_classification\_preprocessor} class in the main repository, \texttt{YAIB}. 

\begin{table}[!htbp]
\caption{Clinical features and units}
\vspace{1em}
\label{tab:features}
\centering
\begin{tabular}{ll}
\toprule
\textbf{Feature} & \textbf{Unit} \\
\midrule
\multicolumn{2}{l}{\textbf{Static}} \\
Age at hospital admission & years \\
Female sex & -- \\
Patient height & cm \\
Patient weight & kg \\
\midrule
\multicolumn{2}{l}{\textbf{Time-varying}} \\
Albumin & g/dL \\
Alkaline phosphatase & IU/L \\
Alanine aminotransferase & IU/L \\
Aspartate aminotransferase & IU/L \\
Band form neutrophils & \% \\
Base excess & mmol/L \\
Bicarbonate & mmol/L \\
Bilirubin (direct) & mg/dL \\
Bilirubin (total) & mg/dL \\
Blood pressure (diastolic) & mmHg \\
Blood pressure (systolic) & mmHg \\
Blood urea nitrogen & mg/dL \\
Calcium & mg/dL \\
Calcium ionized & mmol/L \\
Chloride & mmol/L \\
CO$_2$ partial pressure & mmHg \\
C-reactive protein & mg/L \\
Creatinine & mg/dL \\
Creatine kinase & IU/L \\
Creatine kinase MB & ng/mL \\
Fibrinogen & mg/dL \\
Fraction of inspired oxygen & \% \\
Glucose & mg/dL \\
Haemoglobin & g/dL \\
Heart rate & beats/minute \\
International normalised ratio (INR) & -- \\
Lactate & mmol/L \\
Lymphocytes & \% \\
Magnesium & mg/dL \\
Mean arterial pressure & mmHg \\
Mean cell haemoglobin & pg \\
Mean corpuscular haemoglobin concentration & \% \\
Mean corpuscular volume & fL \\
Methaemoglobin & \% \\
Neutrophils & \% \\
O$_2$ partial pressure & mmHg \\
Oxygen saturation & \% \\
Partial thromboplastin time & sec \\
pH of blood & -- \\
Phosphate & mg/dL \\
Platelets & $1{,}000/\mu$L \\
Potassium & mmol/L \\
Respiratory rate & breaths/minute \\
Sodium & mmol/L \\
Temperature & $^\circ$C \\
Troponin T & ng/mL \\
Urine output & mL \\
White blood cells & $1{,}000/\mu$L \\
\bottomrule
\end{tabular}
\end{table}
\newpage
\subsection{Data distributions and transferability}
\label{Data distributions and transferability}

\begin{figure}[htbp]
  \centering
  \includegraphics[width=0.7\linewidth]{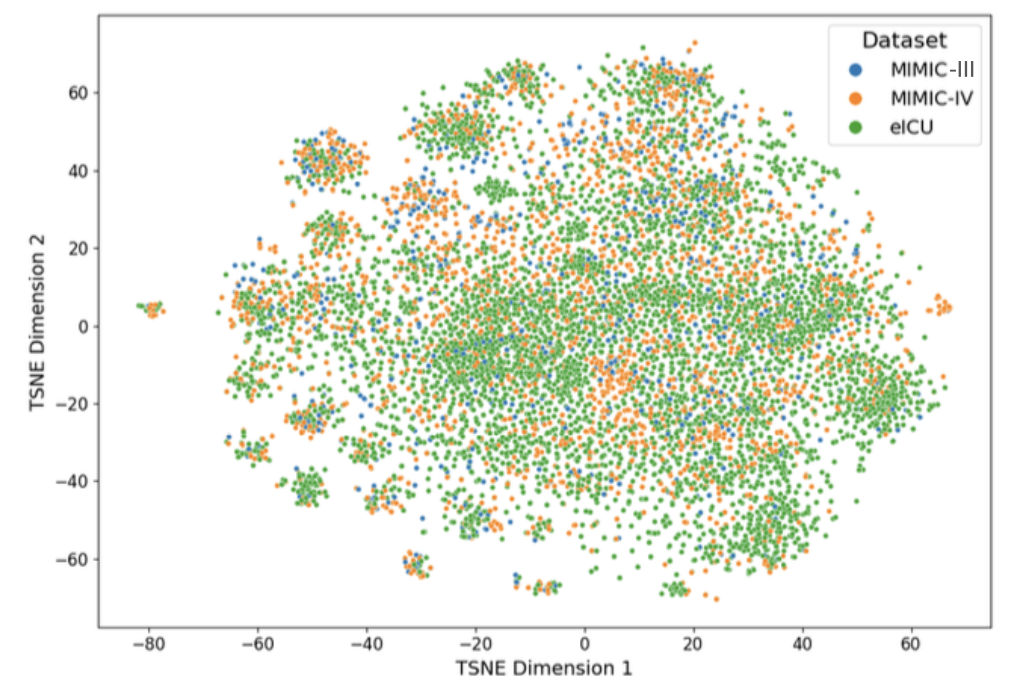}
  \caption[t-SNE of Harmonized and Preprocessed Clinical Datasets]{Two-dimensional t-SNE projection \cite{van2008atsne} of the harmonized and preprocessed datasets used in this study: MIMIC-III, MIMIC-IV, and eICU. Each point represents a time step of a patient stay.}
  \label{tsne}
\end{figure}

\begin{figure}[htbp]
  \centering

  \begin{minipage}[t]{0.45\linewidth}
    \centering
    \includegraphics[width=\linewidth]{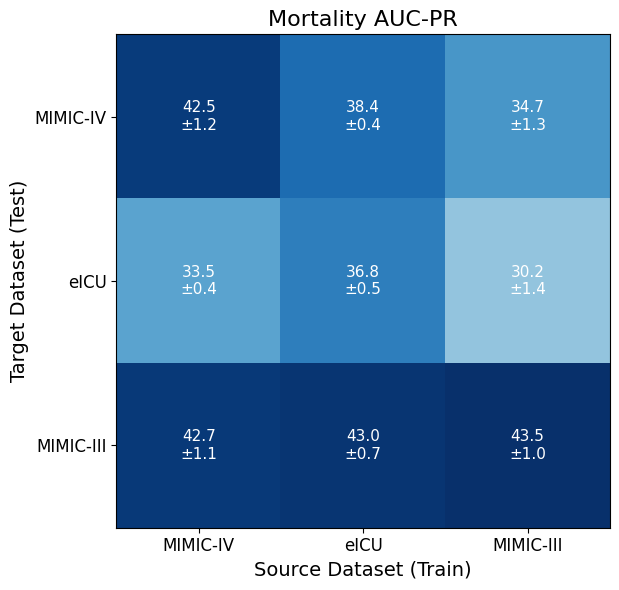}
    \\[0.3em]
    \textbf{\hspace{0.5em}(a)}
  \end{minipage}
  \hfill
  \begin{minipage}[t]{0.45\linewidth}
    \centering
    \includegraphics[width=\linewidth]{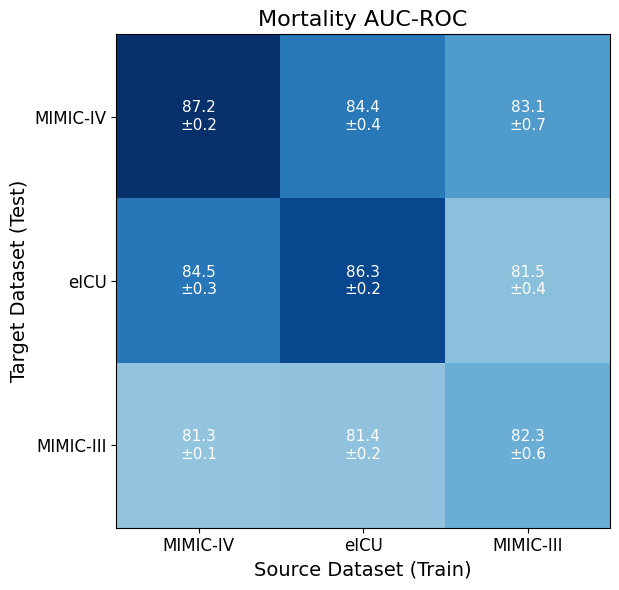}
    \\[0.3em]
    \textbf{\hspace{3em}(b)}
  \end{minipage}

  \vspace{1em}
  \caption[Single-dataset Transfer Performance Heatmap]{Performance of three independently trained models, each evaluated on all three datasets: MIMIC-III, MIMIC-IV, and eICU. \textbf{(a)} AUC-PR scores; \textbf{(b)} AUC-ROC scores. Values reflect mean performance $\pm$ standard deviation on held-out test sets using 5-fold cross-validation.}
  \vspace{1em}
  \label{SCT}
\end{figure}

A t-SNE \cite{van2008atsne} analysis of the preprocessed and harmonized datasets was performed to asses the data distributions of the three datasets before modeling. The results shown in Figure~\ref{tsne}. It reveals that the datasets appear to lie within the same general distribution, which might be a result of similar clinical practices and patient populations.  
BAT was used to train three models, each trained and hyperparameter tuned  on one of the three datasets. The models were then tested independently on all datasets. The models were trained in the 5-fold cross-validation setup described in Appendix~\ref{Data splits}.  The average and standard deviations of AUC-ROC and AUC-PR can be seen in Figure~\ref{SCT}. Each model achieves the best performance on the dataset it was trained on (diagonal). Performance drops relative to the diagonal performance is highest for the model trained on MIMIC-III (AUC-ROC drops between 4.1–4.8) and the lowest for the model trained on MIMIC-IV (AUC-ROC drops between 1.0-1.8) showing different degrees of transferability among the datasets. 

\newpage

\subsection{Hyperparameters}
\label{Hyperparameters}
This appendix provides the optimal hyperparameters used across all experiments. Hyperparameters were selected for each model in two stages. First, a grid search was used to identify a suitable hyperparameter range. Once the hyperparameter ranges were identified, Bayesian hyperparameter optimization (implemented by \citet{yaib_main} in the YAIB framework) was used to select the final hyperparameters. During fine-tuning the models batch size and learning rate were fine-tuned using a grid search while the rest of the hyperparameters were kept the same as during pre-training. 

\newcolumntype{L}{>{\raggedright\arraybackslash}p{0.22\textwidth}} 
\newcolumntype{R}{>{\raggedright\arraybackslash}X}                 

\begin{table}[!htbp]
\caption{Final hyperparameters for pretraining of BAT on pooled datasets MIMIC-III + MIMIC-IV.}
\vspace{1em}
\label{tab:bat_pretrain}
\begin{tabularx}{\textwidth}{@{} p{1.6cm} X @{}}
\toprule
\textbf{Component} & \textbf{Hyperparameters} \\
\midrule
Model & attn\_dropout = 0.207, dropout = 0.364, heads = 1, layers = 2, pooling = max, use\_mask = False, value\_embed\_size = 128 \\
Trainer & batch\_size = 64, epochs = 200, patience = 15, min\_delta = 5e-3 \\
Optimizer & lr = 7.781e-4, weight\_decay = 1e-6 \\
Forecasting & forecast\_horizon = 2, sensors\_count = 48 \\
\bottomrule
\end{tabularx}
\end{table}
\begin{table}[!htbp]
\caption{Final hyperparameters for pretraining of BAT on pooled datasets eICU + MIMIC-III.}
\vspace{1em}
\label{tab:bat_pretrain}
\begin{tabularx}{\textwidth}{@{} p{1.6cm} X @{}}
\toprule
\textbf{Component} & \textbf{Hyperparameters} \\
\midrule
Model & attn\_dropout = 0.357, dropout = 0.249, heads = 1, layers = 8, pooling = max, use\_mask = False, value\_embed\_size = 64 \\
Trainer & batch\_size = 64, epochs = 200, patience = 10, min\_delta = 5e-3 \\
Optimizer & lr = 6.196e-4, weight\_decay = 1e-6 \\
Forecasting & forecast\_horizon = 2, sensors\_count = 48 \\
\bottomrule
\end{tabularx}
\end{table}
\begin{table}[!htbp]
\caption{Final hyperparameters for pretraining of BAT on pooled datasets eICU + MIMIC-IV.}
\vspace{1em}
\label{tab:bat_pretrain}
\begin{tabularx}{\textwidth}{@{} p{1.6cm} X @{}}
\toprule
\textbf{Component} & \textbf{Hyperparameters} \\
\midrule
Model & attn\_dropout = 0.284, dropout = 0.348, heads = 1, layers = 6, pooling = max, use\_mask = False, value\_embed\_size = 64 \\
Trainer & batch\_size = 64, epochs = 200, patience = 10, min\_delta = 5e-3 \\
Optimizer & lr = 3.375e-4, weight\_decay = 1e-6 \\
Forecasting & forecast\_horizon = 2, sensors\_count = 48 \\
\bottomrule
\end{tabularx}
\end{table}
\newcolumntype{L}{>{\raggedright\arraybackslash}p{0.22\textwidth}}
\newcolumntype{R}{>{\raggedright\arraybackslash}X}
\begin{table}[!htbp]
\caption{Final hyperparameters for fine-tuning the classification head of BAT on the datasets MIMIC-III, MIMIC-IV and eICU. All other hyperparameters are kept the same as during pre-training.}
\vspace{1em}
\label{tab:bat_finetune}
\begin{tabularx}{\textwidth}{@{} p{1.6cm} X @{}}
\toprule
\textbf{Dataset} & \textbf{Hyperparameters} \\
\midrule
MIMIC-III & batch\_size = 64, learning\_rate = 1e-2, lr\_scheduler = \mbox{ExponentialLR(gamma = 0.95)} \\
MIMIC-IV  & batch\_size = 64, learning\_rate = 5e-3 , lr\_scheduler = \mbox{ExponentialLR(gamma = 0.95)} \\
eICU      & batch\_size = 24, learning\_rate = 7e-3, lr\_scheduler = \mbox{ExponentialLR(gamma = 0.95)} \\
\bottomrule
\end{tabularx}
\end{table}
\begin{table}[H]
\caption{Final hyperparameters, for fine-tuning the full model, BAT on the datasets MIMIC-III, MIMIC-IV and eICU. All other hyperparameters are kept the same as during pre-training.}
\vspace{1em}
\label{tab:bat_finetune}
\begin{tabularx}{\textwidth}{@{} p{1.6cm} X @{}}
\toprule
\textbf{Dataset} & \textbf{Hyperparameters} \\
\midrule
MIMIC-III & batch\_size = 64, learning\_rate = 9e-05, lr\_scheduler = \mbox{ExponentialLR(gamma = 0.95)} \\
MIMIC-IV  & batch\_size = 24, learning\_rate = 7e-4, lr\_scheduler = \mbox{ExponentialLR(gamma = 0.95)} \\
eICU      & batch\_size = 64, learning\_rate = 5e-4 , lr\_scheduler = \mbox{ExponentialLR(gamma = 0.95)} \\
\bottomrule
\end{tabularx}
\end{table}
\clearpage
\subsection{Data splits}
\label{Data splits}
Models in Figure~\ref{SCT} were trained using a 5-fold cross-validation setup. An initial 80/20 split separated the data into training and test sets, and the training set was further divided into five folds, rotating the validation fold across training iterations. The final performance of the models are reported as the mean ± standard deviation across the five cross-validated models. 
Pre-training of the multi-dataset model was done using the same cross-validation setup. However, only one of the five  models was used for the zero-shot and fine-tuning experiments. This was the one that achieved the lowest masked mean squared error loss during pre-training. The baseline models in the fine-tuning experiment used the same data split as the selected pre-trained model. 
\newpage
\subsection{Models and code availability}
\label{Pre-trained models}
All code related to this project is available at: \url{https://github.com/Katja-Jagd/YAIB} . This repository was originally forked from \url{https://github.com/rvandewater/YAIB} \cite{yaib_main} and has been expanded to support the experiments and methods presented in this work.

\begin{table}[htbp]
\centering
\renewcommand{\arraystretch}{1.2}
\caption{Pre-trained models on combinations of the three datasets of this work (MIMIC-III, MIMIC-IV, and eICU). 
Details on the models, including training and computational resources, are summarized.}
\vspace{1em} 
\label{tab:pretrain_finetune_setup_pivot}
\begin{tabularx}{\textwidth}{l *{3}{X}}
\hline
 & \multicolumn{3}{c}{\textbf{Pre-trained Models}} \\[6pt]
Pooled pre-training datasets
 & \shortstack{MIMIC-IV\\+ eICU} 
 & \shortstack{MIMIC-III\\+ eICU} 
 & \shortstack{MIMIC-III\\+ MIMIC-IV} \\
\hline
Fine-tuning dataset & MIMIC-III & MIMIC-IV & eICU \\
\#Params & 0.86M & 1.13M & 0.97M \\
Pre-training dataset size & 273k & 227k & 100k \\
Pre-training positive class & 6.0\% & 6.26\% & 8.54\% \\
Fine-tuning positive class & 11.9\% & 7.3\% & 5.5\% \\
Pre-training time & $\sim7\mathrm{h}$ & $\sim7\mathrm{h}$ & $\sim3\mathrm{h}$ \\
Pre-training GPU & 1 x A100 (40 GB) & 1 x A100 (40 GB) & 1 x A100 (40 GB) \\
Fine-tuning time & $\leq$ 20 min & $\leq$ 20 min & $\leq$ 20 min \\
Fine-tuning GPU & 1 x V100 (16 GB) & 1 x V100 (16 GB) & 1 x V100 (16 GB) \\
\hline
\end{tabularx}
\end{table}
\newpage
\subsection{Fine-tuning results}
\label{Fine-tuning results}
\begin{table}[!]
\centering
\setlength{\tabcolsep}{6pt}
\renewcommand{\arraystretch}{1.1}
\caption{Model performance, mean AUC-ROC $\pm$ sd , across dataset size ranging from 100 to 9,506 on MIMIC-III, MIMIC-IV, and eICU. The pre-trained BAT models are fine-tuned and the baseline models are trained from scratch on the subsets. Highest performance for each dataset size is indicated in bold, and second highest is underlined.} 
\vspace{1em} 
\label{tab:size-vs-init}

\begin{adjustbox}{width=\linewidth,center}
\begin{tabular}{l r c c c c}
\hline
\noalign{\vskip 2pt}
\shortstack{Fine-tuning/\\Training dataset} & \shortstack{Dataset\\size}
& \shortstack{BAT\\(Fine-tuned full)}
& \shortstack{BAT\\(Fine-tuned head)}
& \shortstack{BAT\\(Scratch)}
& \shortstack{Transformer\\(Scratch)} \\
\hline
          & 100  & 40.88 $\pm$ 0.12 & \textbf{66.14 $\pm$ 5.32} & \underline{57.37 $\pm$ 6.33} & 48.91 $\pm$ 2.23 \\
          & 500  & 35.27 $\pm$ 0.18 & \underline{72.14 $\pm$ 1.64} & \textbf{72.60 $\pm$ 1.91} & 56.52 $\pm$ 7.43 \\
          & 1000 & \textbf{78.30 $\pm$ 0.66} & \underline{74.98 $\pm$ 1.23} & 74.35 $\pm$ 0.89 & 62.83 $\pm$ 7.95 \\
          & 2000 & \textbf{79.72 $\pm$ 0.57} & \underline{78.05 $\pm$ 0.63} & 75.84 $\pm$ 1.98 & 69.99 $\pm$ 0.56 \\
\multicolumn{1}{c}{MIMIC-III} & 3000 & \textbf{80.43 $\pm$ 0.61} & \underline{78.42 $\pm$ 1.15} & 77.73 $\pm$ 0.69 & 71.17 $\pm$ 0.44 \\
          & 5000 & \textbf{81.22 $\pm$ 0.63} & 78.84 $\pm$ 0.55 & \underline{79.77 $\pm$ 0.69} & 71.20 $\pm$ 0.46 \\
          & 7000 & \textbf{81.84 $\pm$ 0.71} & 79.85 $\pm$ 0.31 & \underline{80.10 $\pm$ 0.20} & 72.47 $\pm$ 0.30 \\
          & 9000 & \textbf{82.50 $\pm$ 0.41} & 79.97 $\pm$ 0.35 & \underline{81.12 $\pm$ 0.60} & 72.42 $\pm$ 0.39 \\
          & 9560 & \textbf{82.37 $\pm$ 0.00} & 80.10 $\pm$ 0.00 & \underline{80.73 $\pm$ 0.00} & 72.99 $\pm$ 0.00 \\
\hline
          & 100  & \textbf{58.04 $\pm$ 7.67} & 30.67 $\pm$ 0.19 & \underline{49.62 $\pm$ 11.54} & 47.82 $\pm$ 2.80 \\
          & 500  & \underline{74.85 $\pm$ 3.56} & 70.51 $\pm$ 1.18 & \textbf{75.17 $\pm$ 2.82} & 56.21 $\pm$ 7.67 \\
          & 1000 & \textbf{78.85 $\pm$ 1.11} & 73.52 $\pm$ 2.04 & \underline{77.84 $\pm$ 0.34} & 60.74 $\pm$ 3.52 \\
          & 2000 & \textbf{81.63 $\pm$ 2.01} & 76.57 $\pm$ 0.88 & \underline{79.81 $\pm$ 1.15} & 69.33 $\pm$ 1.24 \\
\multicolumn{1}{c}{MIMIC-IV} & 3000 & \textbf{83.61 $\pm$ 1.24} & 77.87 $\pm$ 1.61 & \underline{81.97 $\pm$ 1.79} & 69.43 $\pm$ 1.74 \\
          & 5000 & \textbf{84.67 $\pm$ 0.81} & 79.43 $\pm$ 0.40 & \underline{83.76 $\pm$ 0.83} & 72.93 $\pm$ 0.66 \\
          & 7000 & \textbf{84.91 $\pm$ 0.93} & 80.16 $\pm$ 0.30 & \underline{84.69 $\pm$ 0.66} & 73.63 $\pm$ 0.64 \\
          & 9000 & \textbf{85.39 $\pm$ 0.59} & 80.60 $\pm$ 0.51 & \underline{84.83 $\pm$ 0.49} & 73.51 $\pm$ 1.30 \\
          & 9560 & \textbf{85.28 $\pm$ 0.80} & 80.70 $\pm$ 0.43 & \underline{85.03 $\pm$ 0.52} & 73.73 $\pm$ 0.77 \\
\hline
          & 100  & 47.87 $\pm$ 0.23 & \underline{64.11 $\pm$ 3.81} & \textbf{65.51 $\pm$ 3.07} & 60.43 $\pm$ 5.30 \\
          & 500  & \textbf{77.80 $\pm$ 2.00} & 71.65 $\pm$ 1.36 & \underline{72.79 $\pm$ 1.68} & 56.09 $\pm$ 4.40 \\
          & 1000 & \textbf{81.42 $\pm$ 0.74} & 74.22 $\pm$ 1.14 & \underline{75.74 $\pm$ 2.08} & 47.93 $\pm$ 11.66 \\
          & 2000 & \textbf{82.99 $\pm$ 1.00} & 76.24 $\pm$ 0.99 & \underline{79.46 $\pm$ 0.98} & 69.51 $\pm$ 0.80 \\
\multicolumn{1}{c}{eICU} & 3000 & \textbf{83.03 $\pm$ 0.50} & 76.68 $\pm$ 0.99 & \underline{81.17 $\pm$ 0.47} & 69.69 $\pm$ 0.30 \\
          & 5000 & \textbf{84.28 $\pm$ 0.33} & 77.89 $\pm$ 1.17 & \underline{82.25 $\pm$ 0.56} & 70.26 $\pm$ 1.02 \\
          & 7000 & \textbf{84.66 $\pm$ 0.32} & 78.38 $\pm$ 0.63 & \underline{83.25 $\pm$ 0.35} & 70.49 $\pm$ 0.81 \\
          & 9000 & \textbf{84.85 $\pm$ 0.58} & 78.80 $\pm$ 1.04 & \underline{83.43 $\pm$ 0.21} & 71.08 $\pm$ 0.78 \\
          & 9506 & \textbf{85.05 $\pm$ 0.25} & 78.26 $\pm$ 1.00 & \underline{83.48 $\pm$ 0.58} & 71.10 $\pm$ 0.61 \\
\hline
\end{tabular}
\end{adjustbox}
\end{table}

\begin{table}[H]
\centering
\setlength{\tabcolsep}{6pt}
\renewcommand{\arraystretch}{1.1}
\caption{Model performance, mean AUC-PR $\pm$ sd , across dataset size ranging from 100 to 9,506 on MIMIC-III, MIMIC-IV, and eICU. The pre-trained BAT models are fine-tuned and the baseline models are trained from scratch on the subsets. Highest performance for each dataset size is indicated in bold, and second highest is underlined.} 
\vspace{1em} 
\label{tab:size-vs-init}

\begin{adjustbox}{width=\linewidth,center}
\begin{tabular}{l r c c c c}
\hline
\noalign{\vskip 2pt}
\shortstack{Fine-tuning/\\Training dataset} & \shortstack{Dataset\\size}
& \shortstack{BAT\\(Fine-tuned full)}
& \shortstack{BAT\\(Fine-tuned head)}
& \shortstack{BAT\\(Scratch)}
& \shortstack{Transformer\\(Scratch)} \\
\hline
          & 100  &  9.73 $\pm$ 0.02 & \textbf{25.77 $\pm$ 5.85} & \underline{15.40 $\pm$ 2.73} & 11.94 $\pm$ 0.76 \\
          & 500  &  8.68 $\pm$ 0.03 & \textbf{30.36 $\pm$ 1.52} & \underline{23.98 $\pm$ 2.62} & 16.75 $\pm$ 4.65 \\
          & 1000 & \textbf{36.24 $\pm$ 1.63} & \underline{33.98 $\pm$ 1.32} & 27.63 $\pm$ 2.23 & 21.34 $\pm$ 4.58 \\
          & 2000 & \textbf{39.01 $\pm$ 1.12} & \underline{37.50 $\pm$ 0.78} & 30.26 $\pm$ 3.07 & 25.41 $\pm$ 0.65 \\
\multicolumn{1}{c}{MIMIC-III} & 3000 & \textbf{40.27 $\pm$ 1.14} & \underline{38.35 $\pm$ 1.62} & 31.92 $\pm$ 1.48 & 26.10 $\pm$ 1.18 \\
          & 5000 & \textbf{41.89 $\pm$ 1.31} & \underline{38.99 $\pm$ 0.96} & 36.30 $\pm$ 0.31 & 26.14 $\pm$ 0.40 \\
          & 7000 & \textbf{42.55 $\pm$ 1.54} & \underline{40.22 $\pm$ 0.54} & 36.32 $\pm$ 1.43 & 26.40 $\pm$ 0.67 \\
          & 9000 & \textbf{43.57 $\pm$ 0.94} & \underline{40.14 $\pm$ 0.70} & 37.09 $\pm$ 1.06 & 27.13 $\pm$ 0.41 \\
          & 9560 & \textbf{43.97 $\pm$ 0.00} & \underline{40.78 $\pm$ 0.00} & 36.73 $\pm$ 0.00 & 27.86 $\pm$ 0.00 \\
\hline
          & 100  & \textbf{12.50 $\pm$ 4.43} &  4.90 $\pm$ 0.02 & \underline{ 8.92 $\pm$ 4.98} &  7.72 $\pm$ 0.66 \\
          & 500  & \textbf{23.96 $\pm$ 3.07} & \underline{23.88 $\pm$ 0.96} & 22.63 $\pm$ 3.23 & 10.92 $\pm$ 2.65 \\
          & 1000 & \textbf{28.98 $\pm$ 0.85} & \underline{26.97 $\pm$ 1.71} & 26.12 $\pm$ 1.95 & 13.06 $\pm$ 1.56 \\
          & 2000 & \textbf{33.19 $\pm$ 2.95} & \underline{28.99 $\pm$ 1.38} & 28.79 $\pm$ 1.64 & 16.47 $\pm$ 0.70 \\
\multicolumn{1}{c}{MIMIC-IV} & 3000 & \textbf{36.60 $\pm$ 1.69} & 30.23 $\pm$ 1.24 & \underline{32.21 $\pm$ 2.63} & 16.86 $\pm$ 0.59 \\
          & 5000 & \textbf{38.10 $\pm$ 1.36} & 31.31 $\pm$ 1.17 & \underline{34.97 $\pm$ 1.03} & 18.00 $\pm$ 1.11 \\
          & 7000 & \textbf{38.60 $\pm$ 1.77} & 31.92 $\pm$ 0.60 & \underline{37.68 $\pm$ 1.25} & 18.65 $\pm$ 0.66 \\
          & 9000 & \textbf{38.75 $\pm$ 0.53} & 32.19 $\pm$ 1.00 & \underline{38.41 $\pm$ 0.88} & 18.91 $\pm$ 1.36 \\
          & 9560 & \textbf{40.05 $\pm$ 0.80} & 32.16 $\pm$ 1.00 & \underline{38.75 $\pm$ 1.84} & 19.62 $\pm$ 0.97 \\
\hline
          & 100  &  7.24 $\pm$ 0.09 & \textbf{14.20 $\pm$ 2.77} & \underline{10.79 $\pm$ 1.83} &  8.82 $\pm$ 1.26 \\
          & 500  & \textbf{23.54 $\pm$ 3.20} & \underline{23.38 $\pm$ 0.98} & 16.78 $\pm$ 2.07 &  8.13 $\pm$ 0.87 \\
          & 1000 & \textbf{28.37 $\pm$ 1.13} & \underline{25.39 $\pm$ 1.56} & 20.86 $\pm$ 3.31 &  6.58 $\pm$ 4.00 \\
          & 2000 & \textbf{31.89 $\pm$ 0.77} & \underline{27.46 $\pm$ 0.50} & 26.12 $\pm$ 0.86 & 13.71 $\pm$ 1.03 \\
\multicolumn{1}{c}{eICU} & 3000 & \textbf{32.32 $\pm$ 1.07} & \underline{28.43 $\pm$ 0.70} & 28.00 $\pm$ 1.38 & 13.90 $\pm$ 0.99 \\
          & 5000 & \textbf{33.89 $\pm$ 0.49} & 29.09 $\pm$ 0.62 & \underline{30.13 $\pm$ 1.37} & 14.41 $\pm$ 0.42 \\
          & 7000 & \textbf{35.14 $\pm$ 0.33} & 29.39 $\pm$ 0.77 & \underline{31.09 $\pm$ 0.98} & 13.98 $\pm$ 0.49 \\
          & 9000 & \textbf{35.20 $\pm$ 0.87} & 29.53 $\pm$ 0.95 & \underline{31.59 $\pm$ 0.94} & 14.61 $\pm$ 1.06 \\
          & 9506 & \textbf{35.17 $\pm$ 0.54} & 29.40 $\pm$ 1.18 & \underline{31.67 $\pm$ 1.25} & 14.18 $\pm$ 1.25 \\
\hline
\end{tabular}
\end{adjustbox}
\end{table}
\end{document}